\documentclass[letterpaper, 10 pt, conference]{ieeeconf}
\IEEEoverridecommandlockouts
\usepackage{amsmath,amssymb,amsfonts}
\usepackage{graphicx}
\usepackage{stfloats} %
\usepackage{booktabs}
\usepackage{siunitx}
\usepackage{xcolor}
\definecolor{revteal}{HTML}{0E8F7E}

\definecolor{linkpink}{HTML}{E0418B}
\newcommand{\acr}[1]{\scalebox{1.15}{#1}}
\usepackage{url}
\usepackage{cite}
\makeatletter
\let\NAT@parse\undefined
\makeatother
\usepackage[draft,unicode,bookmarks=false]{hyperref}
\let\url\nolinkurl
\graphicspath{{figures/}}
  \newcommand{\robot}{MagicBot Z1 humanoid}

\hypersetup{
  pdftitle={Riding the Wave: Sim-to-Real Humanoid Control on Elastic Terrain},
  pdfauthor={},
  pdfsubject={Humanoid locomotion on dynamic elastic terrain},
  pdfkeywords={Humanoid robots; proprioception; reinforcement learning; deformable contact}
}
\ifdefined
\fi
\title{\LARGE \bf
Riding the Wave: Sim-to-Real Humanoid Control on Elastic Terrain
}

\author{%
Yinglei Zhu$^{1,\dagger}$, Qin Li$^{1,\dagger}$, Honghao Pan$^{1,\dagger}$, Bozhao Wang$^{1}$, Binghui Xu$^{2}$, Jie Lu$^{3}$, Wentao Song$^{1}$, \\
Xiyu Wang$^{4}$, Xinyu Zhu$^{5,6}$, Wei Liu$^{5,6}$,
Xinjun Liu$^{1}$, Kai Lu$^{7,*}$, and Huichan Zhao$^{1,*}$%
\thanks{$^{\dagger}$Equal contribution. $^{*}$Corresponding authors.}%
\thanks{$^{1}$Tsinghua University.
$^{2}$National University of Singapore.
$^{3}$East China University of Science and Technology.
$^{4}$Nanjing University.
$^{5}$Magiclab Robotics Technology (Wuxi) Co., Ltd.
$^{6}$Dreame Technology.
$^{7}$University of Oxford.}%
}

\makeatletter
\IEEEaftertitletext{%
  \begin{minipage}{\textwidth}
    \def\@captype{figure}%
    \centering
    \includegraphics[width=\textwidth]{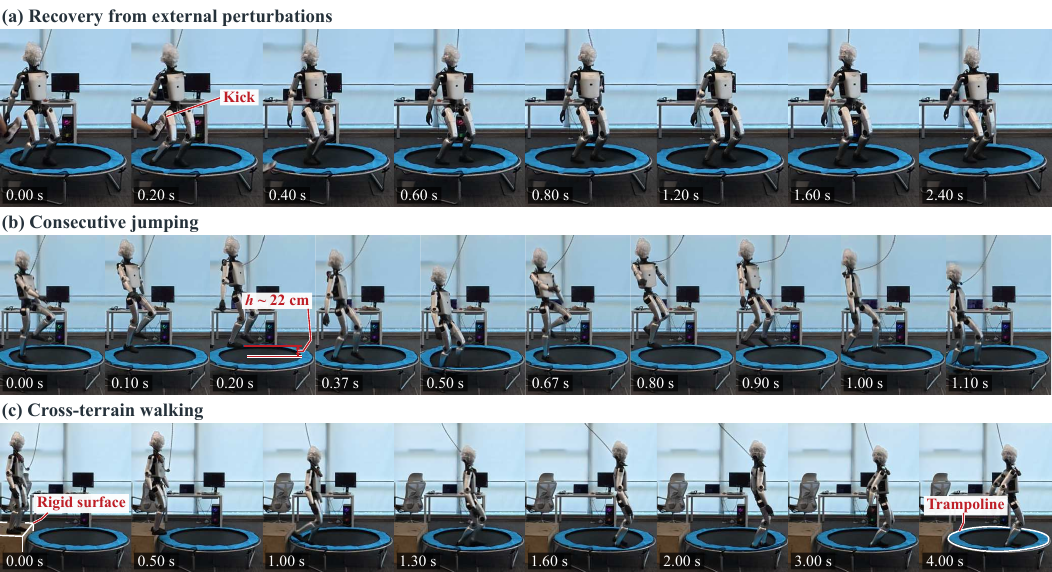}
    \caption{Zero-shot sim-to-real transfer with RIPPLE. On a trampoline, the simulation-trained policies enable a real \robot{} to (a) recover balance after a kick, (b) jump consecutively, and (c) enter from a rigid platform and stabilize after contact. Deployment is zero-shot, using proprioception only with no hardware fine-tuning. Timestamps are relative to the first frame of each sequence.}
    \label{fig:hardware}
  \end{minipage}%
  \vspace{0.5\baselineskip}%
}
\makeatother

\begin{document}
\maketitle
\thispagestyle{plain}
\pagestyle{plain}

\begin{abstract}
Humanoid robots on deformable elastic terrains face a surface whose position and motion are coupled in space and time: deflection and rebound persist between contacts, and load at one foothold alters the support at others, so training must capture how each contact changes the support that follows.
We present RIPPLE (\acr{R}o\-bust \acr{I}n\-ter\-ac\-tion via \acr{P}ro\-pri\-o\-cep\-tive \acr{P}ol\-i\-cies on \acr{L}arge-de\-for\-ma\-tion \acr{E}las\-tic terrain), a simulation-to-real learning framework for humanoid control on such terrains.
RIPPLE integrates ElasticMembrane, a pretensioned mass--spring mesh that models spatially coupled deflection, energy storage, and rebound, together with deformable contact into MJLab for GPU-parallel policy learning.
Staged training then transfers rigid-ground skills to balance and jumping on these terrains using only proprioception, without vision or force sensing.
Deployed zero-shot, the policies enable a \robot{}\footnote{ MagicBot Z1 is a commercially available humanoid robot from MagicLab: \href{https://www.magiclab.top/en/z1}{\nolinkurl{https://www.magiclab.top/en/z1}}.}{} to balance, jump consecutively, and enter a trampoline from a rigid platform.
In simulation, ElasticMembrane training holds standing success of at least 98.7\% (mean over three seeds) on ElasticMembrane, IndependentSprings, and BouncyPlate under 0.3 m/s perturbations, whereas rigid-ground or contact-compliance training transfers poorly to the first two. A fixed policy tolerates stiffness from 0.35 to 2.8 times the training value, beds up to 7 m wide, and seven further simulated support scenarios, so transfer needs no calibration to the trampoline's stiffness.
We plan to release the code soon. The interactive browser demo, videos, and scene configurations are available on the project page:\\
\textcolor{linkpink}{\href{https://fly-pigTH.github.io/wave_rider/}{\mbox{\nolinkurl{https://fly-pigTH.github.io/wave_rider/}}}}
\end{abstract}

\section{Introduction}
{
Humanoid robots are expected to operate in unstructured environments, from disaster response and infrastructure inspection to household assistance. The ground in such settings is often far from an ideal rigid plane: rope bridges, protective nets, floating platforms, mattresses, and inflatable surfaces are compliant and deformable, yielding under load and rebounding when unloaded. Locomotion there means holding balance and tracking commands while the ground itself deforms and moves.

On an elastic terrain, loading changes surface shape and velocity, stores energy, excites motion that persists after unloading, and redistributes force from one foothold to the others. What the robot meets at its next contact therefore depends on the support's evolving state and on its own preceding actions, so a training environment must expose the policy to these dynamics while resolving contact with the moving surface.

Reinforcement learning has enabled agile quadrupedal motion~\cite{hwangbo2019agile,lee2020challenging}, perceptive traversal~\cite{miki2022wild}, and humanoid locomotion from proprioceptive histories~\cite{radosavovic2024humanoid}. Terrain models and training curricula have also enabled locomotion on compliant and yielding surfaces~\cite{choi2023deformable,compliant2024humanoid,mild2025deformable}, where the surface yields locally under the foot. Trampoline-like supports, whose entire surface deflects, stores energy, and keeps moving between footfalls, have only been traversed by a quadruped under model-based control~\cite{wang2020trampoline}; learned humanoid control on such a support has not been demonstrated on hardware.

We present RIPPLE (\acr{R}o\-bust \acr{I}n\-ter\-ac\-tion via \acr{P}ro\-pri\-o\-cep\-tive \acr{P}ol\-i\-cies on \acr{L}arge-de\-for\-ma\-tion \acr{E}las\-tic terrain), a simulation-to-real learning framework for humanoid control on elastic terrains. Its principal model, ElasticMembrane, represents an elastic surface as a mass--spring mesh whose nodes are coupled by axial and diagonal springs. Robot contact excites the membrane, and its subsequent motion feeds back into the robot dynamics. We implement ElasticMembrane in MuJoCo~\cite{todorov2012mujoco} and integrate it into MJLab~\cite{zakka2026mjlab} for parallel reinforcement learning. Policies progress from rigid-ground skills to elastic-terrain balance and jumping through one shared proprioceptive interface, so deployment needs no cameras, foot pressure sensors, or other exteroceptive sensing.

A trampoline realizes local deformation, distributed motion, and rebound in one support. We demonstrate zero-shot deployment on a real \robot{} for balance, consecutive jumping, and entry from a rigid platform (Fig.~\ref{fig:hardware}), against simpler baselines on the same trampoline and against three simpler terrain models --- CompliantContact, IndependentSprings, and BouncyPlate --- in simulation (Sec.~\ref{sec:experiments}).

We introduce humanoid locomotion and dynamic behaviors on large-deformation elastic terrains as a new and challenging setting. To the best of our knowledge, we are the first to study this task with reinforcement learning, and our learned policies generalize across terrain models and transfer zero-shot to a physical trampoline. Our contributions can be summarized in three points:
\begin{enumerate}
  \item A simulation framework built on MuJoCo and MJLab that couples pretensioned ElasticMembrane dynamics with deformable robot--support contact for parallel humanoid policy learning.
  \item A staged training and deployment pipeline that transfers rigid-ground skills to elastic-terrain balance and jumping through a shared proprioceptive interface, with real-trampoline demonstrations.
  \item Generalization experiments that demonstrate policy transfer across multiple terrain models and seven additional simulated scenarios, and quantify performance under changes in stiffness, size, and external disturbances.
\end{enumerate}
\par}

\section{Related Work}
{
\subsection{Modeling Compliant and Deformable Terrain}
Tractable contact models make terrain mechanics accessible to controller design. Romualdi \emph{et al.}~\cite{romualdi2021visco} derive contact forces and torques from a continuum of spring-damper systems and apply the model to humanoid whole-body control in simulation. Vasilopoulos \emph{et al.}~\cite{vasilopoulos2018monopod} incorporate permanent deformation, compaction, and slip into a viscoplastic model for monopod hopping.

For learned locomotion, Choi \emph{et al.}~\cite{choi2023deformable} combine an efficient granular-media model with an adaptive policy that infers terrain properties from interaction, demonstrating quadrupedal motion on sand and other yielding surfaces. Singh \emph{et al.}~\cite{compliant2024humanoid} represent compliance through MuJoCo contact parameters and combine randomized terrain exposure with a curriculum for proprioceptive humanoid walking. MILD~\cite{mild2025deformable} discretizes the foot contact area to model spatially varying granular penetration, enforces consistency across foot contacts, and couples the model to proprioceptive estimation and latent modulation. Together, these works establish terrain representation as a substantive part of policy learning. RIPPLE instead models a bounded elastic terrain whose nodal state carries load transfer and rebound between contacts.

\subsection{Locomotion on Energy-Storing Terrain}
Environmental elasticity introduces both disturbances and opportunities for energy exchange. Wang \emph{et al.}~\cite{wang2020trampoline} demonstrate untethered quadrupedal hopping and bounding on a trampoline, coordinating foot forces to limit pitch oscillations and adjusting leg-extension timing to control hopping height. Divi \emph{et al.}~\cite{divi2023jumping} analyze how latch-mediated actuation influences energy transfer between a small jumper and compliant substrates, with hardware experiments supporting the model. These studies motivate treating support motion and loading history as part of the locomotion problem, but with a quadruped or a small jumper. RIPPLE brings this setting to learned humanoid control on a physical trampoline.

\subsection{Proprioceptive Learning and Sim-to-Real Transfer}
Sim-to-real locomotion requires policies to tolerate differences between simulated and physical dynamics. Dynamics randomization addresses this challenge by varying physical parameters during training~\cite{peng2018dynrand,tan2018simtoreal}. Proprioceptive history also provides information about recent robot--environment interaction: RMA~\cite{kumar2021rma} infers an environment representation from state-action history for its base policy, while Radosavovic \emph{et al.}~\cite{radosavovic2024humanoid} predict humanoid actions with a causal transformer over proprioceptive history.

\subsection{Parallel Simulation for Policy Learning}
Massively parallel simulation accelerates experience collection for locomotion learning~\cite{rudin2022walkminutes}, with Isaac Gym~\cite{makoviychuk2021isaac} and Brax~\cite{freeman2021brax} providing GPU simulation.
MuJoCo~\cite{todorov2012mujoco} provides articulated dynamics, elastic spatial \texttt{tendon}s, and deformable \texttt{flex} contacts.
MuJoCo-Warp~\cite{mujocowarpCode} enables batched GPU execution, while MJLab~\cite{zakka2026mjlab} exposes modular observation, reward, and event interfaces for custom tasks.
RIPPLE adds persistent ElasticMembrane dynamics, deforming contact geometry, and batched resets to this stack.
\par}

\section{RIPPLE Framework}
\label{sec:framework}
\label{sec:model}

\begin{figure*}[!t]
  \centering
  
    \includegraphics[width=\textwidth]{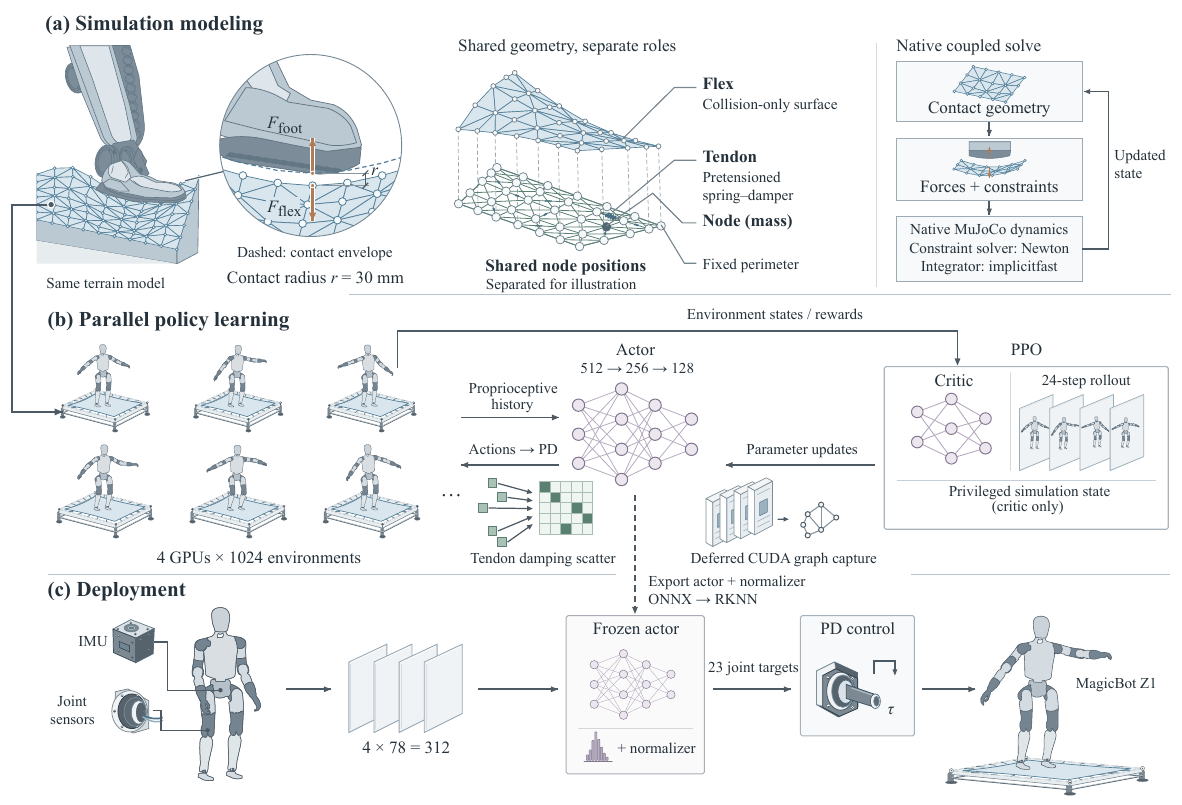}
  
  \caption{{Overview of RIPPLE. (a) Simulation modeling couples mass-carrying nodes, pretensioned {\texttt{tendon}}s, and a collision-only {\texttt{flex}} surface for robot--support interaction. (b) Parallel environments train a proprioceptive actor using PPO and a privileged-state critic. (c) The actor and fixed observation normalizer are exported through ONNX to RKNN for onboard inference.}}
  \label{fig:pipeline}
\end{figure*}

A trampoline combines local indentation, spatial load transfer, and rebound. MuJoCo's continuum-{\texttt{flex}} trampoline example~\mbox{\cite{mujocoTrampolineExample}} exposes no independent pretension parameter, and an unprestressed adaptation deflected excessively in our tests. We therefore construct ElasticMembrane, a configurable pretensioned membrane inspired by thin-shell modeling, and integrate it into RIPPLE with parallel simulation, staged training, and deployment (Fig.~{\ref{fig:pipeline}}).

{
\subsection{Robot--Terrain Mechanics and Simulation Model}
The membrane is discretized as a square grid of point masses (Fig.~\ref{fig:springnet-exploded}). Boundary nodes are fixed; interior nodes translate along three axes. Axial and both diagonal springs couple neighboring nodes.

For nodes at $x_i,x_j$, define $\ell_{ij}=\|x_j-x_i\|$, $e_{ij}=(x_j-x_i)/\ell_{ij}$, and $\dot\ell_{ij}=e_{ij}^\top(\dot x_j-\dot x_i)$. The edge force on node $i$ is
\begin{equation}
 f_{ij}=\big[k_{ij}(\ell_{ij}-\ell_{ij}^0)+c_{ij}\dot\ell_{ij}\big]e_{ij},
 \label{eq:edge}
\end{equation}
with stiffness $k_{ij}$, damping $c_{ij}$, rest length $\ell_{ij}^0$, and opposite force on $j$. Nodes also receive damping $-c_{\mathrm{s}}\dot x_i$. MuJoCo spatial \texttt{tendon}s implement bilateral springs with shortened rest lengths, establishing pretension $T_0$ and elastic potential
\begin{equation}
 \ell_{ij}^0=\ell_{ij}^{\mathrm{ref}}-T_0/k_{ij},\quad
 U_{\mathrm{e}}=\tfrac12\sum_{(i,j)\in\mathcal E}k_{ij}(\ell_{ij}-\ell_{ij}^0)^2.
 \label{eq:elastic}
\end{equation}
Here $\mathcal E$ is the edge set. {For grid spacing $s$, axial and diagonal reference lengths are $s$ and $\sqrt{2}s$. Both use the same stiffness without length scaling and the same pretension $T_0$, which provides out-of-plane restoring forces without a separate bending-energy term.}

{{\textbf{Deformable contact.}} Three components share node positions (Fig.~\ref{fig:springnet-exploded}): bodies carry mass, spatial \texttt{tendon}s provide spring and damping forces, and a collision-only two-dimensional \texttt{flex} mesh defines the contact surface. Each cell has two diagonal spring links and is divided into two contact triangles along one diagonal.}

\begin{figure}[!htbp]
  \centering
  \includegraphics{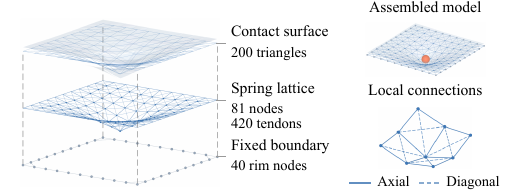} %
  \caption{{ElasticMembrane representation. The exploded view separates the collision-only {\texttt{flex}} surface, pretensioned {\texttt{tendon}} lattice, and fixed boundary for visualization; the surface and lattice share node positions. All views show the same simulated impact state. Solid and dashed lines in the local view denote axial and diagonal springs, respectively.}}
  \label{fig:springnet-exploded}
\end{figure}

{{\textbf{Coupled robot--terrain dynamics.}} For robot generalized position and velocity $(q_{\mathrm{r}},v_{\mathrm{r}})$ and stacked moving-node positions $x$, contact couples the two systems after fixed boundary degrees of freedom are eliminated:}
\begin{equation}
 \begin{aligned}
 M_{\mathrm{r}}(q_{\mathrm{r}})\dot v_{\mathrm{r}}&=S^\top\tau+J_{\mathrm{r}}^\top\lambda-h_{\mathrm{r}}(q_{\mathrm{r}},v_{\mathrm{r}}),\\
 M_{\mathrm{b}}\ddot x&=f_{\mathrm{e}}(x)+f_{\mathrm{d}}(x,\dot x)+M_{\mathrm{b}} g_{\mathrm{b}}+J_{\mathrm{b}}^\top\lambda.
 \end{aligned}\label{eq:coupled}
\end{equation}
{Here $M_{\mathrm{r}},M_{\mathrm{b}}$ are mass matrices, $h_{\mathrm{r}}$ contains robot bias terms, $S$ selects actuated joints, and $\tau$ is torque. The membrane has elastic and damping forces $f_{\mathrm{e}},f_{\mathrm{d}}$ and stacked gravity vectors $g_{\mathrm{b}}$. Signed Jacobians $J_{\mathrm{r}},J_{\mathrm{b}}$ map contact forces $\lambda$ to both systems, while evolving support state $(x,\dot x)$ carries earlier loading into subsequent contacts.}

\textbf{Simulation configuration.} Observed trampoline deflection provided a qualitative reference for selecting the bed parameters. The $1\times1$ m bed uses 10 cm spacing ($11\times11$ nodes) to represent foot-scale deformation. Its 81 interior nodes have a combined nominal mass of approximately 9 kg, setting the membrane's distributed inertia. Link stiffness $k=4$ kN/m controls the change in tension with extension, and pretension $T_0=55$ N supplies the initial out-of-plane stiffness. Edge and nodal damping are each approximately 0.2 N\,s/m, chosen empirically to limit oscillation. A 30 mm \texttt{flex} radius defines the contact envelope; a 12 kg probe dropped at 3 m/s indents the bed by 188 mm and rebounds 361 mm.

The policy runs at 50 Hz with ten 2 ms physics steps per control interval, using the \texttt{implicitfast} integrator and \texttt{Newton} solver~\mbox{\cite{todorov2012mujoco,mujocowarpCode}}. An isolated-bed check showed stable responses at 2 and 3 ms and divergence at 4 ms.

\subsection{Parallel Simulation and Acceleration}
{MuJoCo-Warp executes batched robot--support dynamics on the GPU, while MJLab integrates observations, rewards, and batched resets~\mbox{\cite{mujocowarpCode,zakka2026mjlab}}. Each environment maintains independent robot and support states, and CUDA graph capture follows initialization of the contact buffers.}

{Sparse tendon damping scatter adds $\Delta t\,c_tJ_t^\top J_t$ directly to stored matrix entries corresponding to pairs of nonzero components of the tendon-length Jacobian $J_t$, where $c_t$ is the damping coefficient and $\Delta t$ is the timestep. Together with omitting unused visualization records, this improves simulation throughput by 21.8\%.}\begin{NoHyper}\footnote{{RTX 5090; 1024 parallel ElasticMembrane environments.}}\end{NoHyper}
\par}

{
\subsection{Training Setup}
\label{sec:learning}
\textbf{Policy and control interface.} Training and deployment share a proprioceptive interface for 23 joints: 12 leg joints, waist yaw, and 10 arm joints. The actor observation is
\begin{equation}
 o_t=[\omega_t^{\mathrm{b}},g_t^{\mathrm{b}},c_t,q_{\mathrm{j},t}-q_{\mathrm{j}}^0,\dot q_{\mathrm{j},t},a_{t-1}]
 \in\mathbb R^{78},\label{eq:obs}
\end{equation}
containing body-frame angular velocity, projected unit gravity, three commands, relative joint positions, joint velocities, and previous action. Four oldest-to-newest frames form a normalized 312-dimensional input to an ELU MLP (512, 256, 128). {Only the critic additionally receives base velocity, foot air time, contact-force features, and binary foot-contact indicators.}

{Unclipped actions $a_t\in\mathbb R^{23}$ set joint targets {\mbox{$q_{\mathrm{j},t}^{\mathrm{des}}=q_{\mathrm{j}}^0+0.25a_t$}} in radians. At each physics step, PD control uses the gains and torque limits in Table~{\ref{tab:pd}}.}
\par}

\begin{table}[t]
\centering\scriptsize
\caption{PD gains and nominal joint torque limits.}\label{tab:pd}
\setlength{\tabcolsep}{3pt}
\begin{tabular}{@{}lrrr@{}}
\toprule Joint group & $K_{\mathrm{p}}$ & $K_{\mathrm{d}}$ & $\tau_{\max}$\\\midrule
Hip / knee / waist yaw & 113.06 & 7.20 & 120\\
Ankle pitch / roll & 118.67 & 7.55 & 80/20\\
Shoulder / elbow / wrist yaw & 59.34 & 3.78 & 50\\
\bottomrule
\end{tabular}
\par\vspace{2pt}
{\scriptsize Units: $K_{\mathrm{p}}$: N\,m/rad; $K_{\mathrm{d}}$: N\,m\,s/rad; $\tau_{\max}$: N\,m.\par}
\end{table}

\begin{table}[t]
\centering\scriptsize
\caption{Reward terms of the two training stages.\\
VT: Elastic\-Membrane-VelocityTracking;\\
J: Elastic\-Membrane-Jump.}
\label{tab:rewards}
\setlength{\tabcolsep}{3pt}
\begin{tabular}{@{}llc@{}}
\toprule
Term & Expression & Stage\\
\midrule
Linear velocity & $\exp[-(\lVert c_{v}-v^{\mathrm{b}}_{\mathrm{xy}}\rVert^{2}+(v^{\mathrm{b}}_{\mathrm{z}})^{2})/\sigma_{v}^{2}]$ & VT\\
Angular velocity & $\exp[-((c_{\omega}-\omega^{\mathrm{b}}_{\mathrm{z}})^{2}+\lVert\omega^{\mathrm{b}}_{\mathrm{xy}}\rVert^{2})/\sigma_{\omega}^{2}]$ & VT,\,J\\
Root height & $\exp[-(z_{\mathrm{r}}-z_{\mathrm{r}}^{0})^{2}/\sigma_{h}^{2}]$ & VT\\
Height band & $\max(0,\lvert z_{\mathrm{r}}-z_{\mathrm{r}}^{0}\rvert-\delta_{h})^{2}$ & VT\\
Vertical velocity & $\max(0,\lvert v_{\mathrm{z}}\rvert-\bar v)^{2}$ & VT\\
Jump height & $\exp[-(h_{\mathrm{b}}-h_{0}-c_{\mathrm{h}})^{2}/\sigma_{\mathrm{j}}^{2}]$ & J\\
Takeoff velocity & $\operatorname{clip}(v_{\mathrm{z}},0,\bar v_{\mathrm{z}})$ & J\\
Foot air time & $\textstyle\sum_{f}\mathbf{1}\{\underline t<t^{\mathrm{air}}_{f}<\bar t\}$ & J\\
Stay in place & $\lVert v_{\mathrm{xy}}\rVert^{2}$ & J\\
Yaw-rate error & $(c_{\omega}-\omega^{\mathrm{b}}_{\mathrm{z}})^{2}$ & J\\
Upright & $\exp(-\lVert g^{\mathrm{b}}_{\mathrm{xy}}\rVert^{2}/\sigma_{\mathrm{u}}^{2})$ & VT,\,J\\
Posture & $\exp(-\operatorname{mean}_{j}[(q_{j}-q_{j}^{0})^{2}/\sigma_{j}^{2}])$ & VT,\,J\\
Torso angular rate & $\lVert\omega^{\mathrm{torso}}_{\mathrm{xy}}\rVert^{2}$ & VT,\,J\\
Angular momentum & squared whole-body angular momentum & VT,\,J\\
Joint acceleration & $\lVert\ddot q_{\mathrm{j}}\rVert^{2}$ & VT,\,J\\
Action rate & $\lVert a_{t}-a_{t-1}\rVert^{2}$ & VT,\,J\\
Joint limits & travel outside the soft joint range & VT,\,J\\
Undesired contacts & non-foot contacts above 1 N & VT,\,J\\
Self-collision & substeps with self-contact above a threshold & VT,\,J\\
Termination & $\mathbf{1}\{\text{terminated, excluding timeout}\}$ & VT,\,J\\
\bottomrule
\end{tabular}
\par\vspace{2pt}
\parbox{\columnwidth}{\raggedright\scriptsize Superscript b is the body frame; $c_{v},c_{\omega}$ are the velocity commands and $z_{\mathrm{r}}$ the world-frame root height ($z_{\mathrm{r}}^{0}$ nominal). $\sigma$, $\delta$, and barred symbols are fixed widths and thresholds.}
\end{table}

{
{{\textbf{Staged policy transfer.}} Starting from {pretrained Flat-VelocityTracking}, we train {Elastic\-Membrane-VelocityTracking} and then fine-tune it to obtain {Elastic\-Membrane-Jump}. At each transition, the actor and observation normalizer are inherited from the preceding stage, while the critic and training state are reinitialized. All stages use PPO~\mbox{\cite{schulman2017ppo}} with $4\times1024$ parallel environments on four RTX 5090 GPUs.}

{Elastic\-Membrane-VelocityTracking} assigns half the environments to standing and the remainder to forward/lateral/yaw commands in $[-0.5,1.0]$ m/s, $[-0.5,0.5]$ m/s, and $[-1,1]$ rad/s. {Elastic\-Membrane-Jump} replaces the first command component with a height increment $c_{\mathrm{h}}\in[0.05,0.40]$ m above the nominal root height $h_0=0.70$ m; the root-to-bed height $h_{\mathrm{b}}$ is measured from the nearest node beneath the root plus the contact radius.

\textbf{Reward design.} Table~\ref{tab:rewards} lists the terms of both stages, with weights omitted. Neither stage uses a term specific to elastic terrain, so elastic-terrain behaviors arise from the environment rather than from reward shaping. Relative to velocity tracking, the jump stage drops the linear-velocity, nominal-height, and vertical-velocity terms to permit vertical motion, and adds bed-relative height tracking, takeoff velocity, and foot air time while constraining horizontal drift and yaw. The height term is evaluated at every step rather than only at the apex.

\textbf{Collision detection.} MuJoCo assigns \texttt{flex} contacts to no body, so the body-pair filters that contact sensors normally apply report nothing against the membrane. We therefore detect every robot--support contact from the robot side alone --- foot contact and air time, the critic's contact-force features, and the undesired-contact penalty --- while robot--robot self-collision keeps the usual two-sided filter.

\textbf{Randomization and disturbances.} We randomize foot friction, encoder bias, and base center of mass and add observation noise. Both stages keep ElasticMembrane at its nominal parameters, including 4000 N/m spring stiffness. {Elastic\-Membrane-VelocityTracking} applies root-velocity disturbances, {Elastic\-Membrane-Jump} finite-duration external forces.

\subsection{Sim-to-Real Deployment}
{The actor and fixed observation normalizer are exported through ONNX to RKNN for onboard inference. Deployment reuses the same proprioceptive input and PD interface, with no hardware fine-tuning and no sensing of the support.}
\par}

\section{Experiments}
\label{sec:experiments}
\subsection{{Experimental Setup}}

{
{\textbf{Hardware setup.} The 1.5 m-diameter trampoline is rated for 200 kg; perimeter springs beneath a blue cover connect its black fabric bed to the frame. A steel safety tether remains slack during testing.}

{{\textbf{Training runs and checkpoint selection.}} The deployed Flat-VelocityTracking and Elastic\-Membrane-VelocityTracking policies are the terminal checkpoints of 3000- and 8000-update runs. For Elastic\-Membrane-Jump we deploy the iteration-2400 checkpoint of an 8000-update run, selected for its jump stability and position retention near the trampoline center in simulation.}

{\textbf{Standing-success protocol.} All quantitative standing comparisons use 30 s of zero-command control. After a fixed 20 s wait, horizontal velocity increments of magnitude $u$ are applied at 21, 24, and 27 s. Directions are sampled uniformly over $[0,2\pi)$ with paired seeds across conditions. We score the final 10 s without resetting or excluding trials based on their state after waiting. For samples $\mathcal T_j$ taken every 20 ms from $t_0=20$ s,}
{
\begin{equation}
 \begin{aligned}
 D_j&=\max_{t\in\mathcal T_j}\|p_{\mathrm{xy}}(t)-p_{\mathrm{xy}}(t_0)\|_2,\\
 S_j&=\mathbf{1}\{D_j\leq0.50\,\mathrm{m},\ L_j^{\mathrm{f}}<0.30\,\mathrm{s},\ L_j^{\mathrm{u}}<0.30\,\mathrm{s}\},
 \end{aligned}\label{eq:standing-gate}
\end{equation}
}
{Here $D_j$ measures maximum horizontal displacement from $t_0$ in world coordinates; $L_j^{\mathrm{f}}$ and $L_j^{\mathrm{u}}$ are the longest continuous intervals of falling and loss of bed support. Success rate is the fraction of trials that satisfy $S_j=1$.}

{A fall is detected by excessive tilt ($R_{\mathrm{zz}}<0.7$), pelvis clearance below 0.35 m above local support, or non-foot load-bearing contact above 1 N. Bed support is detected using foot contacts above 1 N or, if absent, five downward rays near the pelvis. Temporary flight is allowed while the bed remains below the robot; the surrounding rigid floor does not count as bed support.}

\begin{figure}[!b]
  \centering
  \includegraphics[width=\columnwidth]{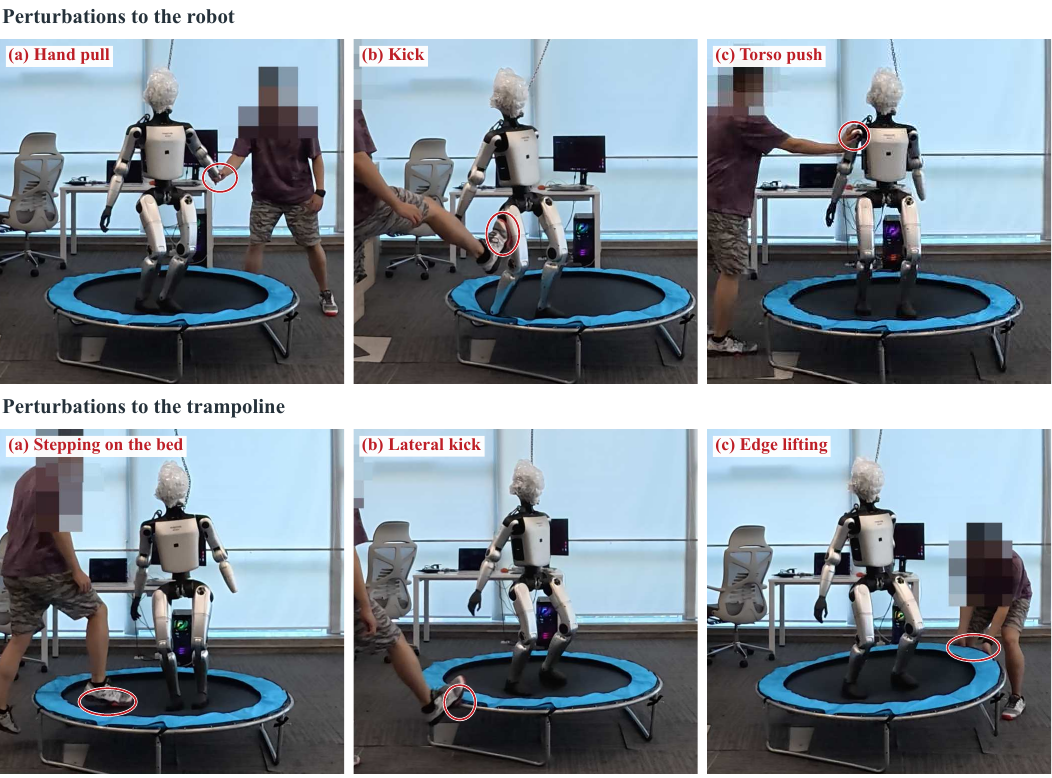}
  \caption{Experimental perturbations during zero-shot hardware deployment. Top row, applied to the robot: (a) pulling its hand, (b) kicking the robot, and (c) pushing its torso. Bottom row, applied to the trampoline: (a) stepping on the bed, (b) kicking the bed laterally, and (c) lifting its edge.}
  \label{fig:perturbation}
\end{figure}

{\textbf{Protocol rationale.} The robot is placed on the support at rest, so the first seconds contain the drop onto the bed, its sag, and free oscillation under the robot's weight. A 20 s wait is sufficient for the robot--support system to settle, so the perturbations test balance on a loaded, settled support rather than recovery from placement; a policy that has fallen or left the bed by then is scored as a failure. The 0.50 m drift radius is half the side length of the simulated 1 m membrane, the distance from bed center to edge.}
\par}

\begin{figure*}[!t]
  \centering
  \includegraphics[width=0.85\textwidth]{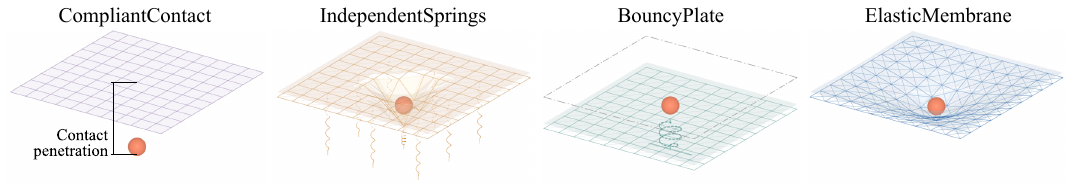}\\[2pt]
  \includegraphics[width=0.85\textwidth]{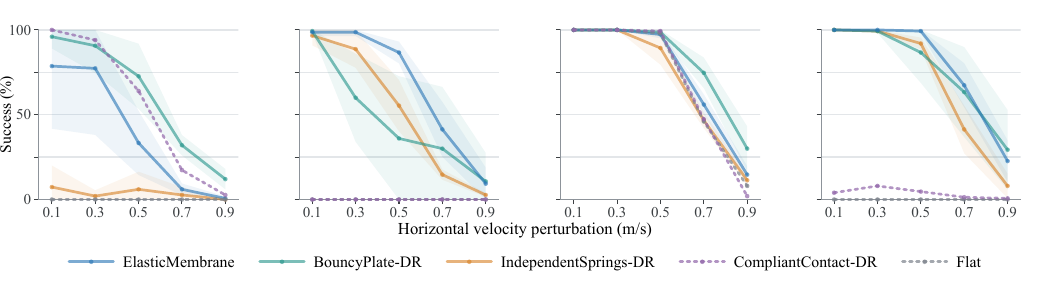}
  \caption{Four terrain models and cross-terrain standing success. Top: responses of four terrain models to the same probe impact: compliant contact against a fixed plane (CompliantContact), independently spring-supported nodes (IndependentSprings), a vertically moving spring-supported rigid plate (BouncyPlate), and a coupled mass--spring mesh (ElasticMembrane). Bottom: standing success on each terrain under 0.1--0.9 m/s velocity perturbations, with a 20 s wait and 0.50 m drift limit, Eq.~\eqref{eq:standing-gate}; curves denote training methods, and colors identify their training terrain. Points show the mean over training seeds 1--3 (Flat-VelocityTracking: one seed); shading is the sample standard deviation across seeds for the three moving-support methods, not a confidence interval (dotted curves: no band); each seed rate uses 50 paired trials.}
  \label{fig:models}
  \label{fig:success-rate}
\end{figure*}

\subsection{{Zero-Shot Hardware Deployment}}

{
Four demonstrations with the RIPPLE policies are documented in the accompanying video:

{
\textbf{Disturbance recovery (Fig.~\ref{fig:hardware}(a), Fig.~\ref{fig:perturbation}).} Under zero-velocity commands, Elastic\-Membrane-VelocityTracking recovers stable standing following external perturbations to the robot and support. These include arm pulls, thigh and torso impacts, and trampoline disturbances induced by stepping, lifting, and kicking.

\textbf{Yaw-command response.} Under yaw-rate commands, Elastic\-Membrane-VelocityTracking turns in the commanded direction, reaching a median 60\% of the commanded rotation. The root-mean-square yaw-rate error is 0.41 rad/s within the training range ($\pm1$ rad/s) and 0.69 rad/s over all active commands.

\textbf{Rigid-to-elastic transition (Fig.~\ref{fig:hardware}(c)).} Under forward-velocity commands, Elastic\-Membrane-VelocityTracking restores trunk stability 2.3 s and 3.3 s after first foot impact in two crossings from the rigid platform onto the trampoline. Recovery requires trunk angular motion to return to its pre-transition walking level for 1 s.

\textbf{Consecutive jumping (Fig.~\ref{fig:hardware}(b)).} Elastic\-Membrane-Jump completes 15 consecutive jumps with flight durations of at least 150 ms and a peak foot clearance of about 22 cm above the bed before stepping onto the perimeter springs.
\par
}

{We also ran the same Elastic\-Membrane-Jump policy on rigid ground; on the trampoline it jumps with lower leg power and mechanical work. Using controller-estimated torques at 500 Hz, we compute total power over the 12 leg joints as $P=\sum_{j=1}^{12}\hat\tau_j\dot q_j$ and positive work as $W_+=\int\max(P,0)\,dt$. The median positive work per landing-to-landing cycle is 39\% lower on the trampoline (59 vs.\ 96 J), and the peak total leg power is 32\% lower (1.33 vs.\ 1.94 kW). Median flight duration rises from 252 to 459 ms and the negative-to-positive work ratio falls from 0.51 to 0.33.}

\begin{table}[t]
\centering\footnotesize
\caption{Zero-shot hardware sessions on the trampoline.}
\label{tab:hardware}
\setlength{\tabcolsep}{2.5pt}
\begin{tabular}{@{}lrll@{}}
\toprule
Training policy & Sim.\ EM & Standing & Entry\\
\midrule
Flat & 0.0\% & $\times$\,9.1 s & $\times$\,0/2\\
CompliantContact & 6.0\% & $\times$\, & $\times$\,\\
Compliant\-Contact-DR & 8.0\% & $\times$\, & $\times$\,\\
IndependentSprings & 38.0\% & $\times$\,14.5 s & $\times$\,\\
Independent\-Springs-DR & 99.3\% & $\times$\,8.4 s & $\times$\,\\
BouncyPlate & 4.0\% & $\times$\,4.4 s & $\times$\,0/1\\
Bouncy\-Plate-DR & 99.3\% & $\times$\,20.8 s & $\checkmark$\,1/1, 36.8 s$^{\ddagger}$\\
ElasticMembrane & 100.0\% & $\checkmark$\,657.5 s, 3.4--4.3$^{\circ\dagger}$ & $\checkmark$\,2/2, 2.3/3.3 s$^{\S}$\\
\bottomrule
\end{tabular}
\par\vspace{2pt}
\parbox{\columnwidth}{\raggedright\scriptsize $\times$: fell; $\checkmark$: completed. Sim.\ EM: simulated standing success on ElasticMembrane at 0.3 m/s (common protocol). Standing: policy-control time; $^{\dagger}$tilt RMS over three perturbed sessions. Entry: rigid-platform crossings; $^{\ddagger}$time on the bed after entry; $^{\S}$trunk settling time in each crossing.}
\end{table}

\textbf{Hardware comparison.} We also deployed the rigid-ground initialization and the six terrain-model baselines (one training seed each) on the same trampoline (Table~\ref{tab:hardware}). Placed on the bed without perturbation, none of the baselines kept stable standing, whereas Elastic\-Membrane-VelocityTracking accumulated 657.5 s of policy control over three sessions, with external perturbations and velocity and yaw commands, without a fall. Entering from the rigid platform, Flat-VelocityTracking and nominal BouncyPlate fell at their first steps onto the bed, Bouncy\-Plate-DR completed its single crossing, and Elastic\-Membrane-VelocityTracking completed both crossings with the trunk settling within 2.3 and 3.3 s. This rapid stabilization after the rigid-to-elastic transition demonstrates the effectiveness of the learned policy on the physical trampoline.
\par}

\subsection{{Effect of Terrain Modeling on Policy Transfer}}

{\textbf{Comparison setup.} Figure~\ref{fig:models} compares four terrain models. CompliantContact uses compliant contact with a fixed plane, following Singh \emph{et al.}~\cite{compliant2024humanoid}. IndependentSprings connects each node to a fixed world anchor through a spring. BouncyPlate translates vertically as a single spring-supported rigid body. ElasticMembrane transmits loads between nodes through a mass--spring mesh. The three moving terrains share the \texttt{flex}-contact construction. All evaluations use the standing-success criterion in Eq.~\eqref{eq:standing-gate}.}

{
The main comparison uses four training methods with three training seeds each: Elastic\-Membrane-VelocityTracking, Independent\-Springs-DR, Bouncy\-Plate-DR, and Compliant\-Contact-DR. All are trained from the same 3000-update Flat-VelocityTracking checkpoint and normalizer with common rewards, commands, observations/actions, generic randomization, and PPO settings, for 8000 updates with $4\times1024$ environments. The -DR variants randomize support stiffness and damping ratio; Compliant\-Contact-DR trains on the CompliantContact plane with independent per-foot contact time constants sampled from $\mathcal U[0.02,0.4]$ s and resampled every 0.3--0.7 s; ElasticMembrane retains fixed support parameters. The Flat-VelocityTracking initialization (one seed) serves as the rigid-ground reference. Terminal checkpoints are evaluated on all four terrains at 0.1, 0.3, 0.5, 0.7, and 0.9 m/s perturbations, with 50 paired trials per condition under the common protocol; results are reported as mean $\pm$ sample standard deviation across training seeds. CompliantContact is an infinite plane with $\tau_{\mathrm{c}}=0.4$ s; the other terrains are finite 1 m beds with rigid surrounding .
\par}

{
\textbf{Necessity of dynamic-support training.} Under the common protocol, the Flat-VelocityTracking reference achieves 0/50 on ElasticMembrane, IndependentSprings, and CompliantContact at 0.3 and 0.5 m/s and succeeds only on BouncyPlate (Fig.~\ref{fig:success-rate}, bottom). Compliant\-Contact-DR, which models compliance only at the contact, transfers to its own plane ($94.0\pm8.7\%$ at 0.3 m/s) and to BouncyPlate, but reaches $8.0\pm13.9\%$ on ElasticMembrane and 0\% on IndependentSprings. In contrast, every policy trained on a support with its own dynamic state exceeds 99\% on ElasticMembrane and BouncyPlate at 0.3 m/s. The deployed checkpoint (training seed 1) reaches 50/50 on all four terrains at 0.3 m/s.

\begin{figure}[t]
\centering
\includegraphics[width=\columnwidth]{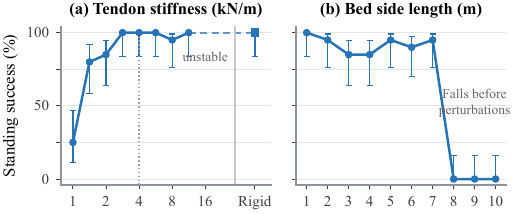}
\caption{Robustness of the deployed Elastic\-Membrane-VelocityTracking policy under the standing-success protocol; each point is 20 paired trials with pointwise 95\% Wilson intervals. (a) Tendon stiffness on the 1 m bed (log scale); the dotted line marks the training value, the square is rigid ground, and the dashed segment bridges the numerically unstable 16 kN/m setting to it. (b) Bed side length at 4000 N/m.}
\label{fig:sensitivity}
\end{figure}

\setcounter{dbltopnumber}{2}
\begin{figure*}[!b]
  \centering
  \includegraphics[width=0.84\textwidth]{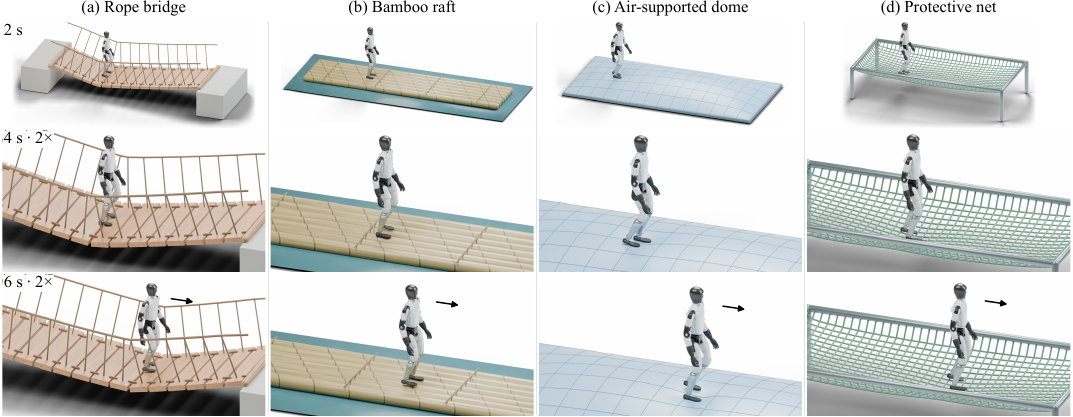}
  \caption{Simulated application sequences at 2, 4, and 6 s (rows). The top row shows overviews; the lower rows are fixed per-column crops at $2\times$. One fixed Elastic\-Membrane-VelocityTracking policy follows a 0.3 m/s forward command (arrows); materials and lighting are illustrative.}
  \label{fig:application-scenes}
\end{figure*}

\textbf{Cross-model transfer.} Among the three dynamic-support methods, Elastic\-Membrane-VelocityTracking has the highest or tied-highest mean success on ElasticMembrane, IndependentSprings, and BouncyPlate at 0.3 and 0.5 m/s (Fig.~\ref{fig:success-rate}). On IndependentSprings at 0.3 m/s, it reaches $98.7\pm2.3\%$, compared with $88.7\pm11.0\%$ for Independent\-Springs-DR and $60.0\pm26.0\%$ for Bouncy\-Plate-DR. On ElasticMembrane at 0.5 m/s, the respective rates are $99.3\pm1.2\%$, $92.0\pm3.5\%$, and $86.7\pm17.9\%$. Bouncy\-Plate-DR is stronger only on the static CompliantContact plane, where it reaches 90.7\% (72\%, 100\%, and 100\% across seeds) at 0.3 m/s versus 77.3\% (100\%, 32\%, and 100\%) for Elastic\-Membrane-VelocityTracking, and Independent\-Springs-DR fails there ($2.0\pm3.5\%$). Conversely, Compliant\-Contact-DR, the only method trained without a moving support, transfers to neither ElasticMembrane nor IndependentSprings in simulation and falls on the physical trampoline (Table~\ref{tab:hardware}). Taken together, ElasticMembrane training yields the highest transfer across the dynamic terrains in simulation and the only policy that kept standing on the trampoline.
\par}

\subsection{{Robustness to Terrain Stiffness and Size}}

\textbf{Stiffness.} We test the deployed Elastic\-Membrane-VelocityTracking policy under the same standing-success protocol with 0.5 m/s perturbations and 20 paired seeds per condition (Fig.~\ref{fig:sensitivity}(a)). On the 1 m bed, tendon stiffness is varied from one quarter to four times the 4000 N/m training value in $\sqrt2$ steps, with pretension scaled to preserve the rest shape. Success is 20/20 at 0.7--1.4$\times$ and 2.8$\times$ the training stiffness, 19/20 at 2$\times$, 16--17/20 at 0.35--0.5$\times$, and 5/20 at 0.25$\times$; on rigid ground the policy passes 20/20. At 4$\times$ the membrane is numerically unstable at the 2 ms timestep, where the bed is already nearly rigid. The fixed policy thus generalizes across stiffness, from 0.35 times the training value to fully rigid ground, without retraining.

{\textbf{Size.} On beds of 1--10 m at 4000 N/m and 0.10 m spacing (Fig.~\ref{fig:sensitivity}(b)), size also raises moving mass and boundary distance. Standing success is 85--100\% on 1--7 m beds and 0/20 at each of 8--10 m; the eleven failures at 2--7 m exceed the drift limit by 0.01--0.17 m without a fall. At 8--10 m every trial falls within the first second after placement, before any perturbation: at fixed stiffness the sag grows with span, from 0.10 m at 1 m to 0.74 m at 7 m and 0.91--1.22 m at 8--10 m, and the drop into it saturates the policy's torques. These falls mark the edge of the operating range at placement, not a loss of push recovery.}

\subsection{{Application Scenarios}}

{
To illustrate applications beyond the trampoline, we deploy the same fixed Elastic\-Membrane-VelocityTracking policy in seven further simulated scenarios whose mechanics differ from a single membrane: rope bridges and protective nets, bamboo rafts and floating foam blankets, spring and inflatable mattresses, and air-supported domes. In each, the unchanged policy stands under a zero command and walks under a 0.3 m/s forward command without falling. Figure~\ref{fig:application-scenes} shows four of the scenes; configurations and full rollouts are provided with our source code and the accompanying video.
\par}

\section{Discussion and Limitations}
\subsection{{Physical Interpretation}}

{\textbf{Elastic response and spatial coupling.} ElasticMembrane transmits loads between nodes and retains motion between contacts, so one model captures the mechanics shared by elastic terrains. Its bounded geometry also acts as an implicit stiffness curriculum: episodes start at the soft bed center and velocity commands carry the robot to the fixed boundary and onto rigid floor, so footholds sweep from compliant to rigid within one episode. The policy thus meets a continuum of effective stiffness without randomizing any terrain parameter (Fig.~\ref{fig:sensitivity}), which accounts for its broad transfer in simulation (Fig.~\ref{fig:success-rate}) and on the trampoline (Table~\ref{tab:hardware}).}

{\textbf{Matched indentation, different rebound.} Matched indentation isolates post-contact response: a 12 kg probe at 3 m/s indents ElasticMembrane, IndependentSprings, and BouncyPlate by 188, 188, and 189 mm but rebounds 361, 304, and 148 mm; on CompliantContact ($\tau_{\mathrm{c}}=0.4$ s) it sinks 436 mm into a fixed plane.}

Policies learn only what their training terrain exposes: without a moving support they fail on ElasticMembrane and IndependentSprings, whereas ElasticMembrane, with the largest rebound and spatial load transfer, transfers most broadly. Transfer still varies across seeds, and Bouncy\-Plate-DR is stronger on the static compliant plane, so the specific contribution of spatial coupling remains to be isolated.

\subsection{{Limitations and Future Work}}
Quantitative validation against the trampoline's load--deflection and rebound responses remains future work; the stiffness sweep (Fig.~\ref{fig:sensitivity}) nevertheless shows that zero-shot transfer needed no fine calibration to its stiffness. Future work will also validate RIPPLE on other morphologies, restructure the MuJoCo-Warp computation for higher throughput, and carry the method beyond the trampoline to physical instances of the simulated scenarios.

\section{Conclusion}
We presented RIPPLE, a sim-to-real framework integrating the ElasticMembrane terrain model with parallel simulation and staged policy training.
Its proprioceptive policies transfer zero-shot to a real humanoid for balance recovery, consecutive jumping, and transitions from a rigid platform onto a trampoline.
RIPPLE is released as open-source infrastructure for further research on legged locomotion over deformable terrain, toward robots that truly ride the wave.

\bibliographystyle{IEEEtran}
\bibliography{bib/refs}

\begin{thebibliography}{10}
\providecommand{\url}[1]{#1}
\csname url@samestyle\endcsname
\providecommand{\newblock}{\relax}
\providecommand{\bibinfo}[2]{#2}
\providecommand{\BIBentrySTDinterwordspacing}{\spaceskip=0pt\relax}
\providecommand{\BIBentryALTinterwordstretchfactor}{4}
\providecommand{\BIBentryALTinterwordspacing}{\spaceskip=\fontdimen2\font plus
\BIBentryALTinterwordstretchfactor\fontdimen3\font minus \fontdimen4\font\relax}
\providecommand{\BIBforeignlanguage}[2]{{%
\expandafter\ifx\csname l@#1\endcsname\relax
\typeout{** WARNING: IEEEtran.bst: No hyphenation pattern has been}%
\typeout{** loaded for the language `#1'. Using the pattern for}%
\typeout{** the default language instead.}%
\else
\language=\csname l@#1\endcsname
\fi
#2}}
\providecommand{\BIBdecl}{\relax}
\BIBdecl

\bibitem{hwangbo2019agile}
J.~Hwangbo, J.~Lee \emph{et~al.}, ``Learning agile and dynamic motor skills for legged robots,'' \emph{Sci. Robot.}, vol.~4, no.~26, 2019, {A}rt. no. eaau5872.

\bibitem{lee2020challenging}
J.~Lee, J.~Hwangbo \emph{et~al.}, ``Learning quadrupedal locomotion over challenging terrain,'' \emph{Sci. Robot.}, vol.~5, no.~47, 2020, {A}rt. no. eabc5986.

\bibitem{miki2022wild}
T.~Miki, J.~Lee \emph{et~al.}, ``Learning robust perceptive locomotion for quadrupedal robots in the wild,'' \emph{Sci. Robot.}, vol.~7, no.~62, 2022, {A}rt. no. eabk2822.

\bibitem{radosavovic2024humanoid}
I.~Radosavovic, T.~Xiao \emph{et~al.}, ``Real-world humanoid locomotion with reinforcement learning,'' \emph{Sci. Robot.}, vol.~9, no.~89, 2024, {A}rt. no. eadi9579.

\bibitem{choi2023deformable}
S.~Choi, G.~Ji \emph{et~al.}, ``Learning quadrupedal locomotion on deformable terrain,'' \emph{Sci. Robot.}, vol.~8, no.~74, 2023, {A}rt. no. eade2256.

\bibitem{compliant2024humanoid}
R.~P. Singh, M.~Morisawa \emph{et~al.}, ``Robust humanoid walking on compliant and uneven terrain with deep reinforcement learning,'' in \emph{IEEE-RAS Int. Conf. Humanoid Robots (Humanoids)}, 2024, pp. 497--504.

\bibitem{mild2025deformable}
Z.~Luo, J.~Zhang \emph{et~al.}, ``{MILD}: Tractable terrain modeling for learning improved bipedal locomotion on deformable surfaces,'' \emph{IEEE Robot. Autom. Lett.}, vol.~11, no.~2, pp. 1922--1929, 2026.

\bibitem{wang2020trampoline}
B.~Wang, C.~Zhou \emph{et~al.}, ``Untethered quadrupedal hopping and bounding on a trampoline,'' \emph{Front. Mech. Eng.}, vol.~15, no.~2, pp. 181--192, 2020.

\bibitem{todorov2012mujoco}
E.~Todorov, T.~Erez \emph{et~al.}, ``{MuJoCo}: A physics engine for model-based control,'' in \emph{IEEE/RSJ Int. Conf. Intelligent Robots and Systems (IROS)}, 2012, pp. 5026--5033.

\bibitem{zakka2026mjlab}
K.~Zakka, Q.~Liao \emph{et~al.}, ``{mjlab}: A lightweight framework for {GPU}-accelerated robot learning,'' \emph{arXiv preprint arXiv:2601.22074}, 2026.

\bibitem{romualdi2021visco}
G.~Romualdi, S.~Dafarra \emph{et~al.}, ``Modeling of visco-elastic environments for humanoid robot motion control,'' \emph{IEEE Robot. Autom. Lett.}, vol.~6, no.~3, pp. 4289--4296, 2021.

\bibitem{vasilopoulos2018monopod}
V.~Vasilopoulos, I.~S. Paraskevas \emph{et~al.}, ``Monopod hopping on compliant terrains,'' \emph{Robot. Auton. Syst.}, vol. 102, pp. 13--26, 2018.

\bibitem{divi2023jumping}
S.~Divi, C.~Reynaga \emph{et~al.}, ``Adapting small jumping robots to compliant environments,'' \emph{J. R. Soc. Interface}, vol.~20, no. 200, 2023, {A}rt. no. 20220778.

\bibitem{peng2018dynrand}
X.~B. Peng, M.~Andrychowicz \emph{et~al.}, ``Sim-to-real transfer of robotic control with dynamics randomization,'' in \emph{IEEE Int. Conf. Robotics and Automation (ICRA)}, 2018, pp. 3803--3810.

\bibitem{tan2018simtoreal}
J.~Tan, T.~Zhang \emph{et~al.}, ``Sim-to-real: Learning agile locomotion for quadruped robots,'' in \emph{Robotics: Science and Systems (RSS)}, 2018.

\bibitem{kumar2021rma}
A.~Kumar, Z.~Fu \emph{et~al.}, ``{RMA}: Rapid motor adaptation for legged robots,'' in \emph{Robotics: Science and Systems (RSS)}, 2021.

\bibitem{rudin2022walkminutes}
N.~Rudin, D.~Hoeller \emph{et~al.}, ``Learning to walk in minutes using massively parallel deep reinforcement learning,'' in \emph{Conf. Robot Learning (CoRL)}, vol. 164, 2022, pp. 91--100.

\bibitem{makoviychuk2021isaac}
V.~Makoviychuk, L.~Wawrzyniak \emph{et~al.}, ``{Isaac Gym}: High performance {GPU}-based physics simulation for robot learning,'' \emph{arXiv preprint arXiv:2108.10470}, 2021.

\bibitem{freeman2021brax}
C.~D. Freeman, E.~Frey \emph{et~al.}, ``{Brax}: A differentiable physics engine for large scale rigid body simulation,'' \emph{arXiv preprint arXiv:2106.13281}, 2021.

\bibitem{mujocowarpCode}
\BIBentryALTinterwordspacing
{Google DeepMind} and {NVIDIA}, ``{MuJoCo Warp (MJWarp)},'' 2026. [Online]. Available: \url{https://github.com/google-deepmind/mujoco_warp}
\BIBentrySTDinterwordspacing

\bibitem{mujocoTrampolineExample}
\BIBentryALTinterwordspacing
{Google DeepMind}, ``{MuJoCo} trampoline example.'' [Online]. Available: \url{https://github.com/google-deepmind/mujoco}
\BIBentrySTDinterwordspacing

\bibitem{schulman2017ppo}
J.~Schulman, F.~Wolski \emph{et~al.}, ``Proximal policy optimization algorithms,'' \emph{arXiv preprint arXiv:1707.06347}, 2017.

\end{thebibliography}
\end{document}